%% file: naaclhlt2019.tex
\documentclass[11pt,a4paper]{article}
\usepackage[hyperref]{naaclhlt2019}
\usepackage{times}
\usepackage{latexsym}
\usepackage{multirow}\usepackage{graphicx} 
\usepackage{subfig} 
\usepackage{textcomp}

\usepackage{url}

\aclfinalcopy 

\title{Improving Natural Language Interaction with Robots Using Advice}

\author{Nikhil Mehta \\
  Department of Computer Science \\
  Purdue University \\
  West Lafayette, IN 47907 \\
  {\tt mehta52@purdue.edu} \\\And
  Dan Goldwasser \\
  Department of Computer Science \\
  Purdue University \\
  West Lafayette, IN 47907 \\
  {\tt dgoldwas@purdue.edu} \\}

\date{}
\begin{document}
\maketitle
\begin{abstract}
Over the last few years, there has been growing interest in learning models for physically grounded language understanding tasks, such as the popular blocks world domain. These works typically view this problem as a single-step process, in which a human operator gives an instruction and an automated agent is evaluated on its ability to execute it. In this paper we take the first step towards increasing the bandwidth of this interaction, and suggest a protocol for including advice, high-level observations about the task, which can help constrain the agent’s prediction. We evaluate our approach on the blocks world task, and show that even simple advice can help lead to significant performance improvements. To help reduce the effort involved in supplying the advice, we also explore model self-generated advice which can still improve results.
\end{abstract}

\section{Introduction}
\label{sec:intro}
\input{introduction.tex}

 \section{Model}
 \label{sec:model}
\input{model.tex}
 
 \section{Experiments}
 \label{sec:experiments}
 \input{experiments.tex}
 
 \section{Summary}
 \label{sec:summary}
 \input{summary.tex}
 
 \section*{Acknowledgements}
 \label{sec:acknowledgements}
 \input{acknowledgements.tex}
 
\bibliography{naaclhlt2019}
\bibliographystyle{acl_natbib}

\newpage

\appendix

\section{Supplemental Material}
\label{sec:supplemental}
In this section, we provide implementation details for our models. The dataset we use has 2,493 training, 360 development, and 720 test examples.

\subsection{Advice Sentence Generation}
\label{sec:advice-sent-gen}
Our advice sentences are generated by filling in appropriate regions/directions into varying sentences. For example, given an advice sentence placeholder, \textit{The target is in the \underline{\hspace{1cm}}}, and a coordinate in the lower left, we would generate the restrictive advice sentence: \textit{The target is in the lower left.} At test time, we use variations of this sentence such as: \textit{The block's region is the lower left}, to avoid memorization. We showed in Table~\ref{table:resultsTable} the importance of our pre-trained models in enabling sentence variability and advice understanding ($\mathtt{M4}$ vs $\mathtt{M5}$).

\subsection{Pre-trained Model Details}
\label{sec:pre-trained-details}
All of our pre-trained models are trained on random coordinates and advice sentences. 

The model used to comprehend restrictive advice is trained as a binary prediction problem, and must output a positive prediction if the random input coordinate is in the region described by the input advice sentence. This design allows the model to understand the meaning of the advice sentence by determining if the random coordinate follows the sentence. 

Our architecture takes as input an advice sentence $ s = w_1, w_2, ..., w_n $, passes it through a trained embedding layer of size 100, a LSTM of size 256, and outputs the hidden state representations $ \{h_n\} $. The last hidden state $ h_n $ is embedded using a Fully Connected (FC) layer of size 100. Each axis of a random input coordinate $(x, y, z) $ is also passed into the network and embedded using a FC layer of size 100. These 4 FC layers are summed up and passed through a final FC layer $ O $ of size 2, which is then followed by a softmax. All FC layers use the RELU activation function. We train this as a binary prediction problem using cross-entropy loss, the Adam optimizer, and a learning rate of 0.001. Gradient clipping \cite{pascanu2013difficulty} is used on the LSTM parameters to avoid exploding gradients.

The model for corrective advice is identical to the one for restrictive advice, except the final FC layer $ O $ has size 3, and the model is trained to output a coordinate that follows the advice. If it outputs an advice-following coordinate, the model receives 0 loss. Otherwise a mean square regression loss is provided, where the ground truth is some random coordinate that does follow the advice. 

\subsection{End-to-end Advice Model Details}
\label{sec:end-to-end-details}
The end-to-end model is trained and tested on the training and test split from \cite{bisk2016natural}. We load and freeze the LSTM, embedding layer, LSTM hidden state $h_n$ and the FC layer following it from the pre-trained model into the end-to-end model. This last FC layer is passed through a FC layer of size 256, which is then summed with the LSTM hidden state of the original \citeauthor{bisk2016natural} model. We are able to accurately generate and feed-in the advice to the pre-trained portion of this model (just like a human would), as we have access to the true source/target coordinates from the dataset. The rest of the architecture and training procedure is identical to \citeauthor{bisk2016natural}.

\subsection{Model Advice Generation Details}
\label{sec:model-advice-gen-details}

For the model that self-generates the advice (Figure~\ref{fig:advice_generation}), we use a neural network model, passing the instruction from the \citeauthor{bisk2016natural} dataset into an embedding layer of size 256, followed by a LSTM of size 256. We then embed the blocks-world grid using a FC layer of size 20 (the maximum number of blocks there are), final FC layer of size 4 (when dividing the grid into 4 regions), and softmax with cross-entropy loss. We train using Adam optimizer \cite{kingma2014adam} and a learning rate of 0.0001. All FC layers use the Leaky RELU activation function \cite{maas2013rectifier}. We note that if we change the final FC layer to size 3 and train this model as a coordinate prediction problem, the performance is worse than \citeauthor{bisk2016natural}. This shows that the overall performance improvement shown in Section~\ref{sec:advice-gen} is due to considering self-generated advice.

When using the model self-generated advice in the end-to-end model, we use accurate advice at training time (computed using the train split of \citeauthor{bisk2016natural}), and the self-generated advice at test time. The self-generated test advice is generated by training the advice generation model on the \citeauthor{bisk2016natural} dataset, and using the best performing model (evaluated on the dev split) to generate advice for the test data. Thus, no human interaction is needed. 

When using retry advice, we also use accurate advice at training time and self-generated advice at test time. However, we generate two sets of self-generated advice, one for the most confident region prediction, and another for the next most confident one (determined by the softmax scores and explained in Section~\ref{sec:retry-advice}). If the general region of the coordinate prediction in the first iteration of running the end-to-end model (with the most confident self-generated advice) is incorrect, the human operator will provide retry advice, and we then feed in the second most confident advice (using that for the final prediction). 

Input-specific self-generated advice is generated in two iterations. In the first iteration, we run the trained \citeauthor{bisk2016natural} model on the test split, and then create an advice region (of the same size as when we divided the board into four quadrants) centered at the predicted coordinate (but not exceeding the boundary of the board). In the second iteration, this advice is fed into the \citeauthor{bisk2016natural} model just like Section~\ref{sec:restrictive-advice}. Thus, again, no human interaction is needed.
\end{document}

%% file: introduction.tex
The problem of constructing an artificial agent capable of understanding and executing human instructions is one of the oldest long-standing AI challenges~\cite{winograd1972understanding}. This problem has numerous applications in  various domains (planning, navigation and assembly) and can help accommodate seamless interaction with personal assistants in many environments. Due to its central role in AI and wide applicability, this problem has seen a surge of interest recently~\cite{macmahon2006walk,branavan2009reinforcement,chen2011learning,tellex2011understanding,matuszek2012joint,kim2013adapting,misra2017mapping}.

Recent works~\cite{bisk2016natural,tan2018source} focus on exploring deep learning methods for grounding spatial language. In this popular setup, human communication with robots is viewed as a single-step process, in which a natural language (NL) instruction is provided, and an outcome is observed.

Our goal in this paper is to explore different approaches for relaxing the \textit{single step} assumption, and present initial results which we hope would motivate future work in this direction. Similar to interactive dialog systems~\cite{allen1995trains,rybski2007interactive,wen2017network}, we view this problem as an \textit{interactive} process, in which the human operator can observe the agents' response to their instruction and adjust it by providing advice, a form of online feedback. Specifically, the advice consists of a short sentence, simplifying the user's intent. We utilize two types of advice, one restricting the agent's search space to a general region (\textit{restrictive advice}), and the other telling the agent the appropriate direction (up, down, left, right) to adjust its current prediction (\textit{corrective advice}). 

\begin{figure}[t!]
  \centering
  \includegraphics[scale=0.19]{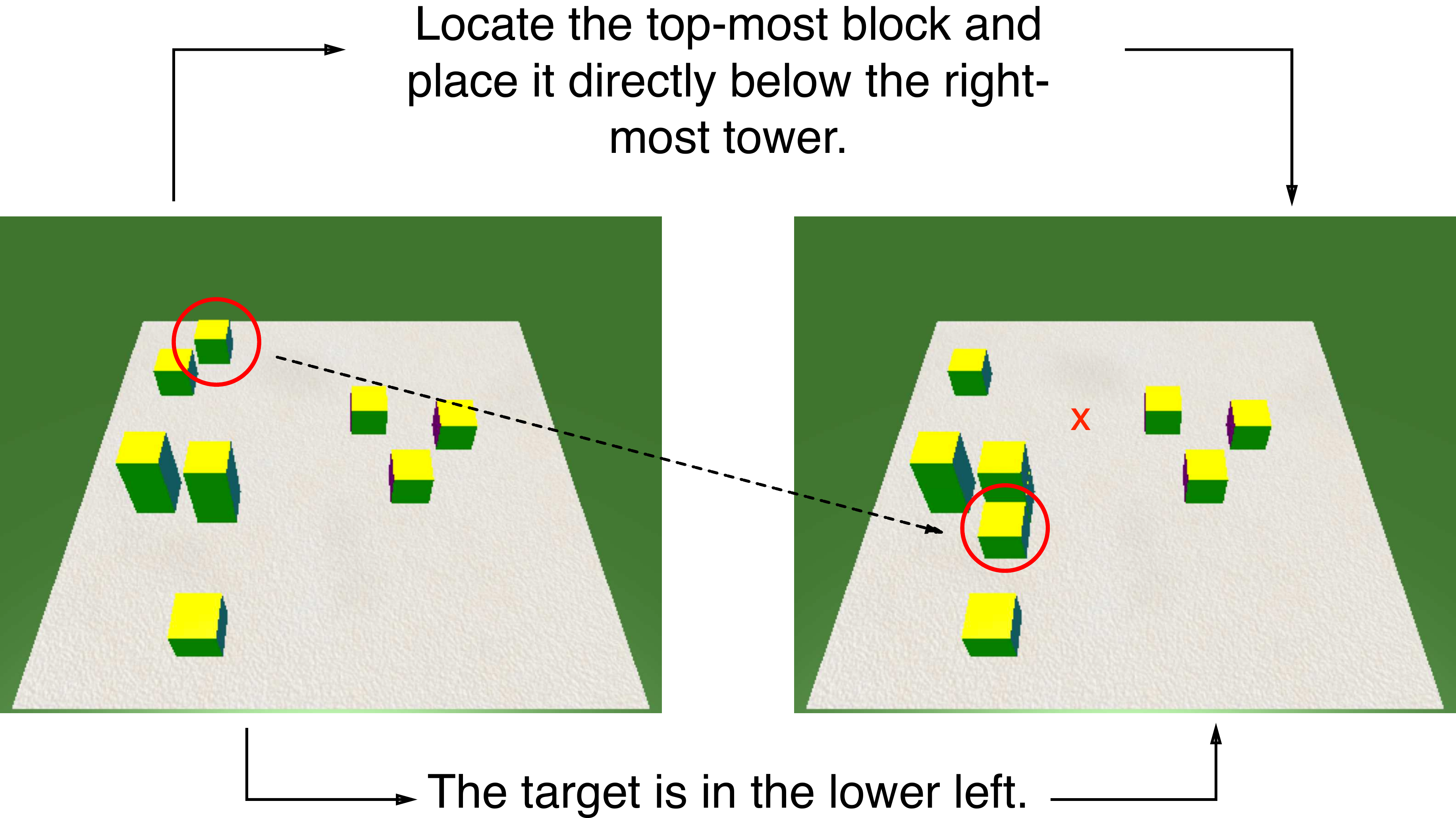}
  \caption{Based on the instruction (upper sentence) the model predicts the coordinates of the block and its target location. The `x' represents an incorrect prediction, corrected by the provided advice (lower sentence).}
  \label{fig:exampleAdvice}
  \vspace{-15pt}
\end{figure}

Our focus is on the challenging task of moving blocks on a grid~\cite{winograd1972understanding}, in which the agent is given only an instruction and the state of the grid, and must predict the coordinates of where a block must be moved. We follow the difficult experimental settings suggested by~\cite{bisk2016natural}, in which the blocks are unlabeled and can only be referenced by their spatial properties. Fig.~\ref{fig:exampleAdvice} describes our settings and uses the advice \textit{``the target is in the lower left''}, to restrict the agents search space after observing  the incorrect prediction placed the target block in the top half of the board.

\begin{figure*}[ht!]
   \centering
   \subfloat[\label{genworkflow}]{%
      \includegraphics[scale=0.39]{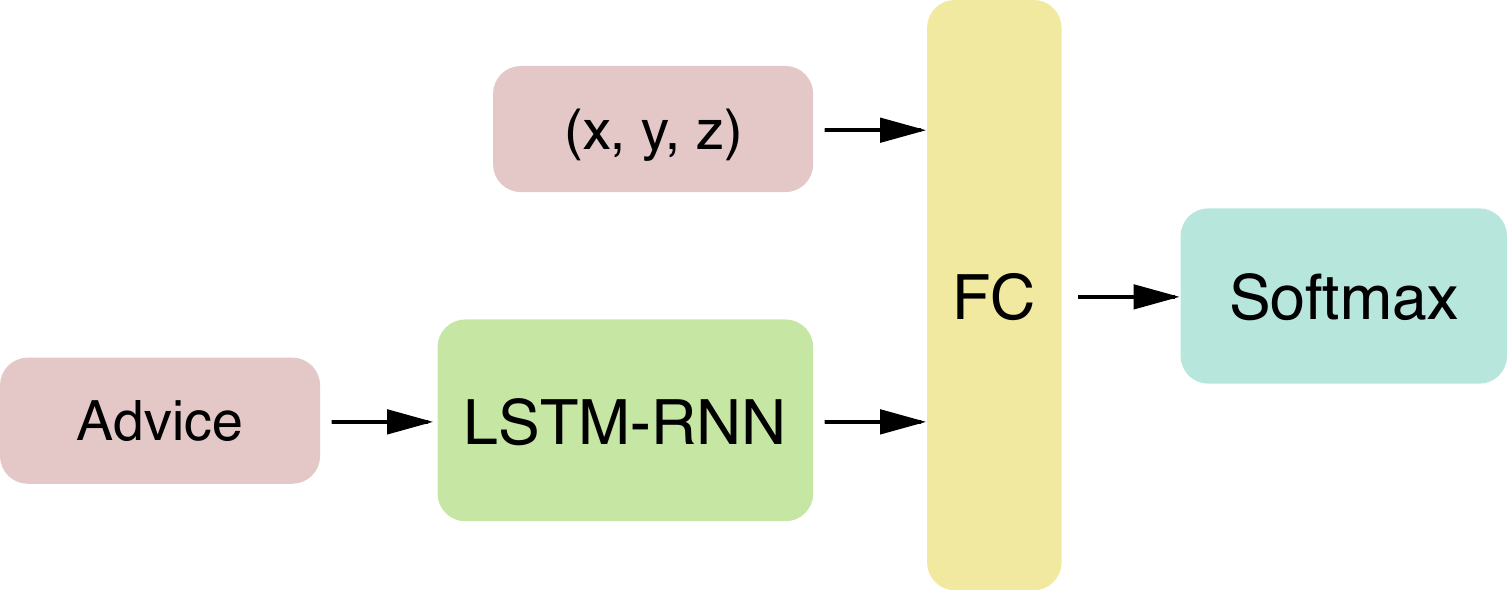}}
\hfill
   \subfloat[\label{pyramidprocess} ]{%
      \includegraphics[scale=0.39]{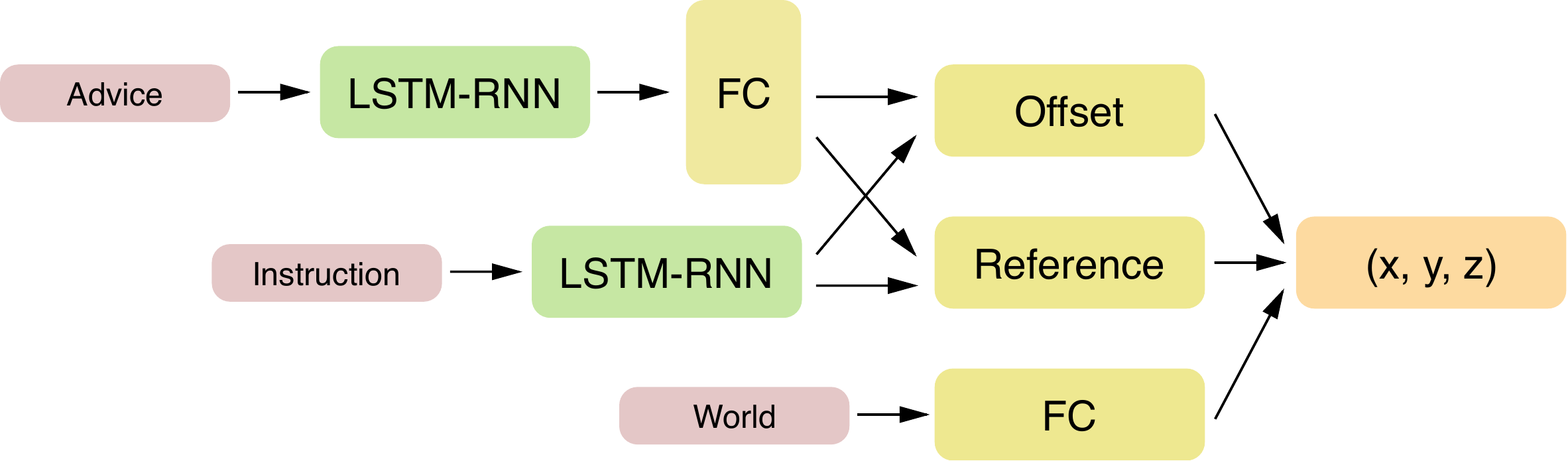}}
\caption{Our advice architectures. (a) Pre-Trained Advice Understanding Model. (b) \cite{bisk2016natural} End-to-End architecture with our pre-trained model. World represents the board state, while offset and reference represent fully connected layers used to identify the offset and reference blocks.}
\label{fig:architecture}
\end{figure*}

To accommodate these settings, we take a two step approach. First, we ground the advice text in the simulated blocks-world environment by training a neural network. In the second step, we integrate the trained advice network into the end-to-end neural model proposed by \citeauthor{bisk2016natural} Our architecture is described in Fig.~\ref{fig:architecture}. The experiments we run show that this end-to-end advice model successfully grounds the meaning of our advice.

We propose four novel interactive advice-based protocols that can be applied on any robot communication architecture, ordered in terms of decreasing human effort. As expected, as human effort lessens, performance does worsen, but all protocols outperform \citeauthor{bisk2016natural} (whom our model is \textit{identical} to besides the inclusion of advice).

Most notably, we explore the notion of model \textit{self-generated} advice, which significantly reduces/eliminates human effort. In this approach, a model is trained to automatically generate restrictive advice for a given scenario, based on the assumption that it is easier to predict a region containing the target coordinates rather than their exact location. We validate this assumption by developing a neural architecture to predict the restrictive advice and show it can help improve the overall prediction quality, despite having no human assistance.

%% file: model.tex
This section describes the architecture we developed for understanding advice, and how to incorporate it into the original \citeauthor{bisk2016natural} model to make better predictions. We begin by defining the Blocks World task and the types of advice we use. We then introduce a model for grounding the advice and a method for incorporating the pre-trained advice understanding module into the original model. Finally, we discuss an architecture for \textit{advice generation}, a method for self-predicting the advice to avoid any human intervention. Further details of our models and advice generation process are in the Appendix.

\subsection{Blocks World Task Definition}
\label{sec:task-def}
Given an input state, consisting of all the block positions on a board, and a NL instruction, the model has to predict the coordinates of the source block to be moved and its target location. We follow the definition by \cite{bisk2016natural} and due to space constraints refer the reader to that paper.

\subsection{Advice}
\label{sec:advice}
The two types of advice we devise in this paper are designed to assist the prediction agent by providing simpler instructions in addition to the original input. The first, restrictive advice, informs the agent about the general region of the source / target coordinates, such as \textit{top left}. These regions are determined by dividing the board into equally sized sections (two halves, four quadrants). The second type of advice, corrective advice, observes the agents\textquotesingle {} predictions and determines which direction (up, down, left, right) they must be adjusted to get closer to the target. Both of these are representative of information a human could easily provide to a robot in various ways (speech, using assisted devices, etc.), to help correct its predictions. Specific examples are shown below.
\begin{table}[h!]
\begin{center}
\begin{tabular}{|c|c|c|}
  \hline
  {\textbf{Predicted}} & {\textbf{Target}} & {\textbf{Advice}}\\
 \hline
  - & (-0.5, 0.5, 0.5) & In the top left.\\
 (-0.5, 0.5, 0.9) & (-0.5, 0.5, 0.5) & Move down. \\
 \hline
\end{tabular}
\end{center}
\vspace{-20pt}
\end{table}


\begin{table*}[t!]
\begin{center}
\begin{tabular}{|p{8.6cm}|c|c|c|c|}
  \hline
  \multirow{2}{8.6cm}{\textbf{Model}} & \multicolumn{2}{c|}{\textbf{Source}} & \multicolumn{2}{c|}{\textbf{Target}}\\
 \cline{2-5}
 & \textbf{Median} & \textbf{Mean} & \textbf{Median} & \textbf{Mean}\\
 \hline
$\mathtt{M1:}$ \citeauthor{bisk2016natural} & 3.29 & 3.47 & 3.60 & 3.70\\
 $\mathtt{M2:}$ Our Replication of \citeauthor{bisk2016natural} & 3.13 & 3.42 & 3.29 & 3.50\\
  $\mathtt{M3:}$ \citeauthor{tan2018source} & -- & 2.21 & 2.78 & 3.07\\
 \hline
 $\mathtt {M4:}$ Restrictive Advice w/o Pre-Trained Model & 3.88 & 3.83 & 3.56 & 3.43 \\ 
 \hline
  $\mathtt {M5:}$ 4 Regions Restrictive Advice & 2.23 & 2.21 & 2.18 & 2.19 \\
 \hline
 $\mathtt {M6:}$ Corrective Advice & 2.76 & 2.94 & 2.72 & 3.06 \\
 \hline
 $ \mathtt{M7:}$ 4 Regions Retry Advice & 2.41 & 3.02 & 2.42 & 3.14\\
 \hline
 $ \mathtt{M8:}$ 2 Regions Model Self-Generated Advice & 3.01 & 3.31 & 3.08 & 3.36 \\ 
 $ \mathtt{M9:}$ Input-Specific Model Self-Generated Advice & 2.87 & 3.12 & 2.99 & 3.26 \\
 \hline
\end{tabular}
\end{center}
\caption{Results for our models compared to previous models evaluated as distance from gold prediction normalized by block length for source and target coordinate prediction.}
\label{table:resultsTable}
\vspace{-15pt}
\end{table*}

\begin{figure}[h!]
  \centering
  \includegraphics[scale=0.30]{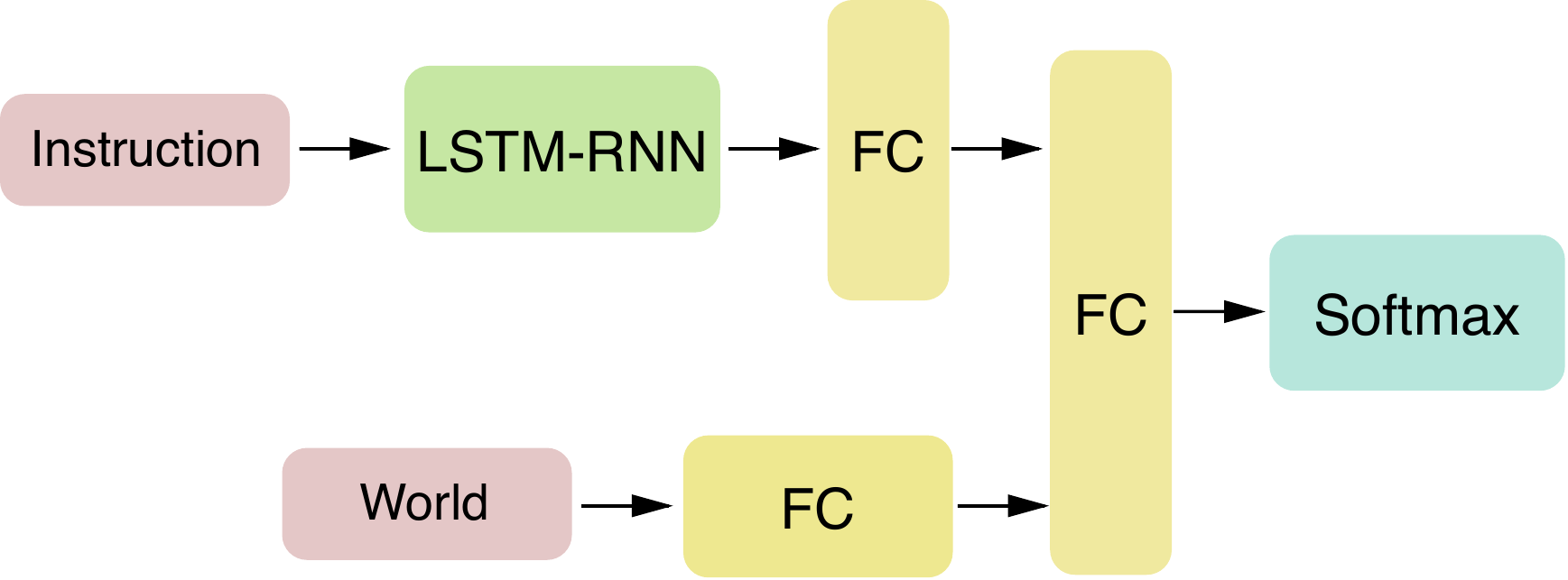}
  \caption{Model for Generating Advice.}
  \label{fig:advice_generation}
\end{figure}

\subsection{Advice Grounding}
\label{sec:grounding}
We pre-train a neural network model to accurately understand the advice. For both types of advice, a LSTM-RNN \cite{hochreiter1997long} is used to read the advice sentence $ s = w_1, w_2, ..., w_n $ and output the hidden state representations $ \{h_n\} $. Prior to this, a word embedding layer is used to project the input words into high-dimension vectors $ \{w_i\} $. 

For restrictive advice, the last state from the LSTM $ h_n $ is fed along with a random coordinate into a Fully Connected (FC) layer. The network must output a positive prediction if the random coordinate is in the region described by the advice sentence, and negative otherwise. 

For corrective advice, the last  state from the LSTM $ h_n $ is fed along with a random coordinate into a FC layer, and the network must output a coordinate that follows the advice. For example, if the advice is \textit{move the block down}, the predicted coordinate must be below the random input coordinate. If the advice is followed, the network receives 0 loss, otherwise a MSE regression loss.

\subsection{End-to-End Training}
\label{sec:end-to-end}
The pre-trained model from Section~\ref{sec:grounding} that understands various advice text is incorporated into the best performing End-to-End RNN architecture proposed in \cite{bisk2016natural} by adding a FC layer to the pre-trained LSTM state $ h_n $ and summing it with the LSTM hidden state of the original model (as shown in Figure~\ref{fig:architecture}b). We load and freeze the best performing parameters from our pre-trained model into the relevant portion of this end-to-end architecture, and train it on the original task of predicting the coordinates of the source / target location, with the addition of advice input.

\subsection{Advice Generation}
\label{sec:self-generated}

We use a neural network model to self-generate restrictive advice (as shown in Figure~\ref{fig:advice_generation}), passing the instruction into an embedding layer followed by a LSTM, the board state into a FC layer, concatenating these into a FC layer, and finally using a softmax to classify the input example into a region. We train this architecture and then run it on the test set, generate the appropriate advice based on the region the data is classified in, and use that as test advice input for the end-to-end architecture from section~\ref{sec:end-to-end}.

%% file: experiments.tex
Next, we present our experiments over our four different advice protocols, each with decreasing human effort and overall performance. In each protocol, we provide advice to the end-to-end model from Section~\ref{sec:end-to-end}, whether it is given by a human user or model self-generated. Our results, evaluated on each model's mean and median prediction error, are presented in Table~\ref{table:resultsTable}. We always compare to the baseline \citeauthor{bisk2016natural} model, which our model is identical to besides the addition of advice (and we always beat), and the state-of-the-art best non-ensemble \citeauthor{tan2018source} architecture. Note that \citeauthor{tan2018source} use an advanced neural architecture and a different training procedure (source prediction trained as classification). We hypothesize that using the advice mechanism over this more complex architecture would lead to further improvements, and leave it for future work.

The pre-trained advice grounding models from Section~\ref{sec:grounding} achieve 99.99\% accuracy, and are vital, as shown by the poor performance without them ($\mathtt{M4}$ vs $\mathtt{M5}$). These grounding models allow the end-to-end architecture to generalize to the variability in advice utterances.

\subsection{Restrictive Advice}
\label{sec:restrictive-advice}
When training the end-to-end model from Section~\ref{sec:end-to-end}, we provide restrictive advice at training time for only half the examples. For every epoch, a different half set of examples (determined randomly) receive advice. This mechanism gives the model a chance to learn to interpret each example with and without advice, so that it can handle the interactivity without overfitting to one setup. This setup also gave the best performance.

At test time, the advice is provided only whenever the predictions fall in the wrong general region, just like a human would. As seen in Table~\ref{table:resultsTable}, this model ($\mathtt{M5}$) significantly outperforms both baselines ($\mathtt{M1}$, $\mathtt{M3}$). We note that the performance did not improve much when advice was always provided, showing that this model was able to perform well in its absence and does not rely on it (due to our choice not to provide advice all the time in training). In fact a human would only have to provide restrictive advice for 395/720 examples, and the model always follows it.\footnote{We note that the performance does not improve from \citeauthor{bisk2016natural} if advice is only provided at train time.}

\subsection{Corrective Advice}
\label{sec:corrective-advice}
We train corrective advice identically to restrictive advice from Section~\ref{sec:restrictive-advice}, except we train in two separate iterations. This is necessary as the model must learn to adjust its predictions based on the advice, which is why it is first trained to make the normal prediction (first iteration), then trained to adjust the prediction (second iteration).

In the first iteration, we train identically to \cite{bisk2016natural} with no advice, but in the second iteration corrective advice is generated based on which direction the predictions must be adjusted to be more accurate. This case is simpler than restrictive advice, since the human operator just has to provide the direction to adjust the predictions, rather than the precise region of the coordinates. However, the performance does worsen ($\mathtt{M5}$ vs $\mathtt{M6}$).

\subsection{Retry Advice}
\label{sec:retry-advice}
In Section~\ref{sec:self-generated}, we introduced a model that was able to self-generate restrictive advice by predicting the general region of the block coordinates given the NL instruction and blocks world. Table~\ref{table:accuracyGenAdvice} shows this model's accuracy on that task when the board is split into 4 regions. As this is a hard problem with low accuracy ($\mathtt{A1}$), we instead generate advice for the top 2 most confident predictions (determined by the softmax scores) ($\mathtt{A2}$).

We now introduce a new multi-step \textit{retry} advice protocol. In the first step, the model from Section~\ref{sec:self-generated} self-generates restrictive advice based on the most confident predicted region, which it uses as input in the end-to-end model. If the user believes the coordinate prediction based on this advice is wrong, it can tell the model to ``retry'', and then the second most likely restrictive advice will be used. Thus, the only human feedback needed now is telling the model to ``retry'', rather than accurate advice as before. The performance of this ($\mathtt{M7}$) still significantly outperforms \citeauthor{bisk2016natural} and is close to \citeauthor{tan2018source} on target prediction. 

\begin{table}[h!]
\begin{center}
\begin{tabular}{|p{3.8cm}|c|c|}
  \hline
  {\textbf{Regions}} & {\textbf{Source}} & {\textbf{Target}}\\
 \hline
 $\mathtt{A1:}$ 4 & 47\% & 40\% \\
 $\mathtt{A2:}$ 4, Top 2 Confidence & 73\% & 70\% \\
 $\mathtt{A3:}$ 4, Input-Specific & 67\% & 62\% \\
 \hline
\end{tabular}
\end{center}
\caption{Accuracy of model self-generated advice.}
\label{table:accuracyGenAdvice}
\end{table}

\subsection{Model Self-Advice Generation}
\label{sec:advice-gen}

We now aim to avoid any human interaction, by letting the model completely self-generate the advice. Accomplishing it would allow us to improve the model's performance without additional human effort. We experimented with two approaches. In the first, we generate advice as described in Section~\ref{sec:retry-advice}. However, instead of having the user ask the model to ``retry'', we treat the top 2 confidence regions as a general region, and provide that as advice input as described in Section~\ref{sec:restrictive-advice}. In this case, there is a performance improvement over \citeauthor{bisk2016natural} with no human effort required ($\mathtt{M8}$ in Table~\ref{table:resultsTable}). 

\begin{figure}[t!]
  \centering
  \subfloat[\label{genworkflow}]{%
      \includegraphics[scale=0.22]{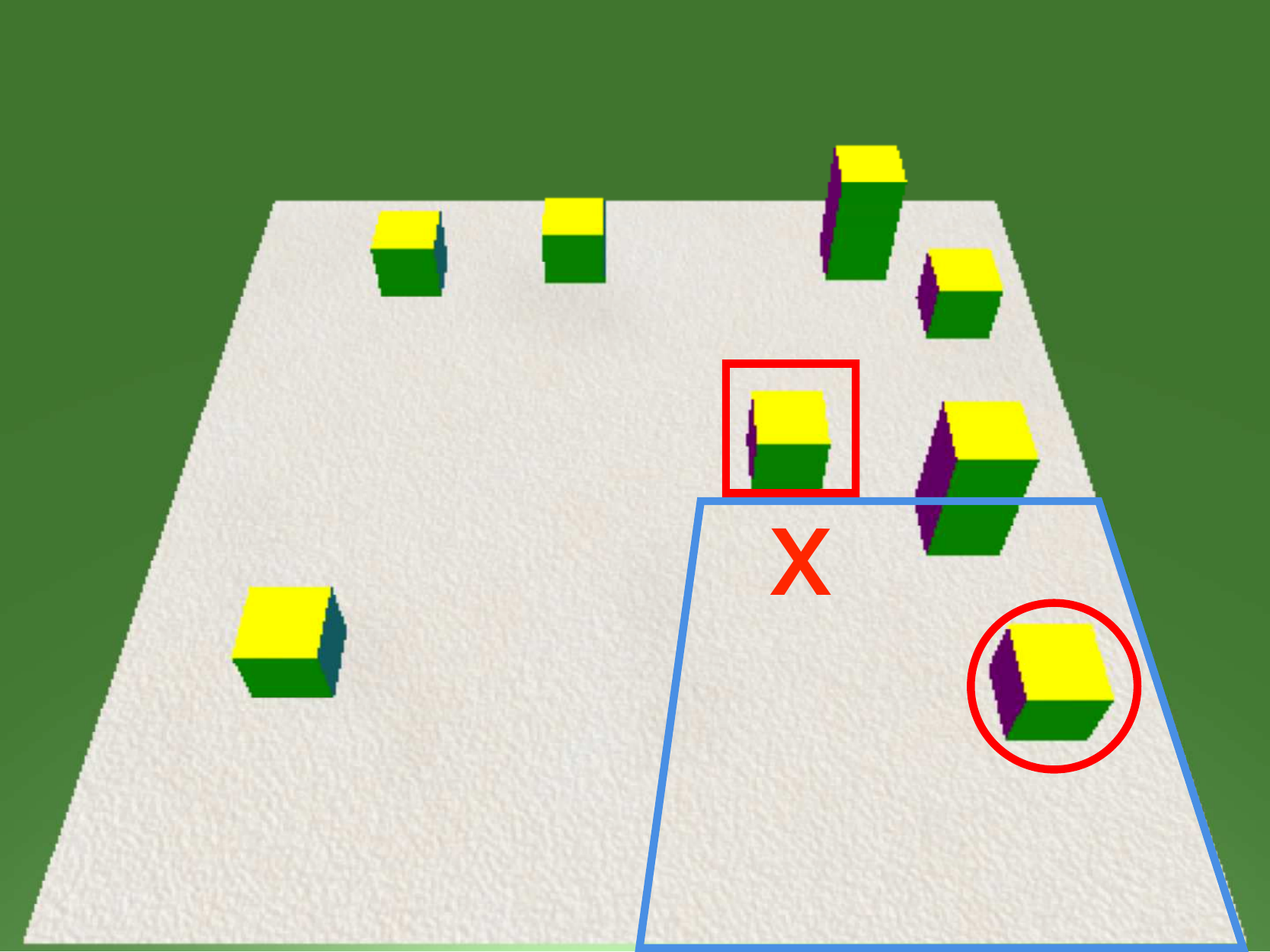}}
\hfill
  \subfloat[\label{pyramidprocess} ]{%
      \includegraphics[scale=0.22]{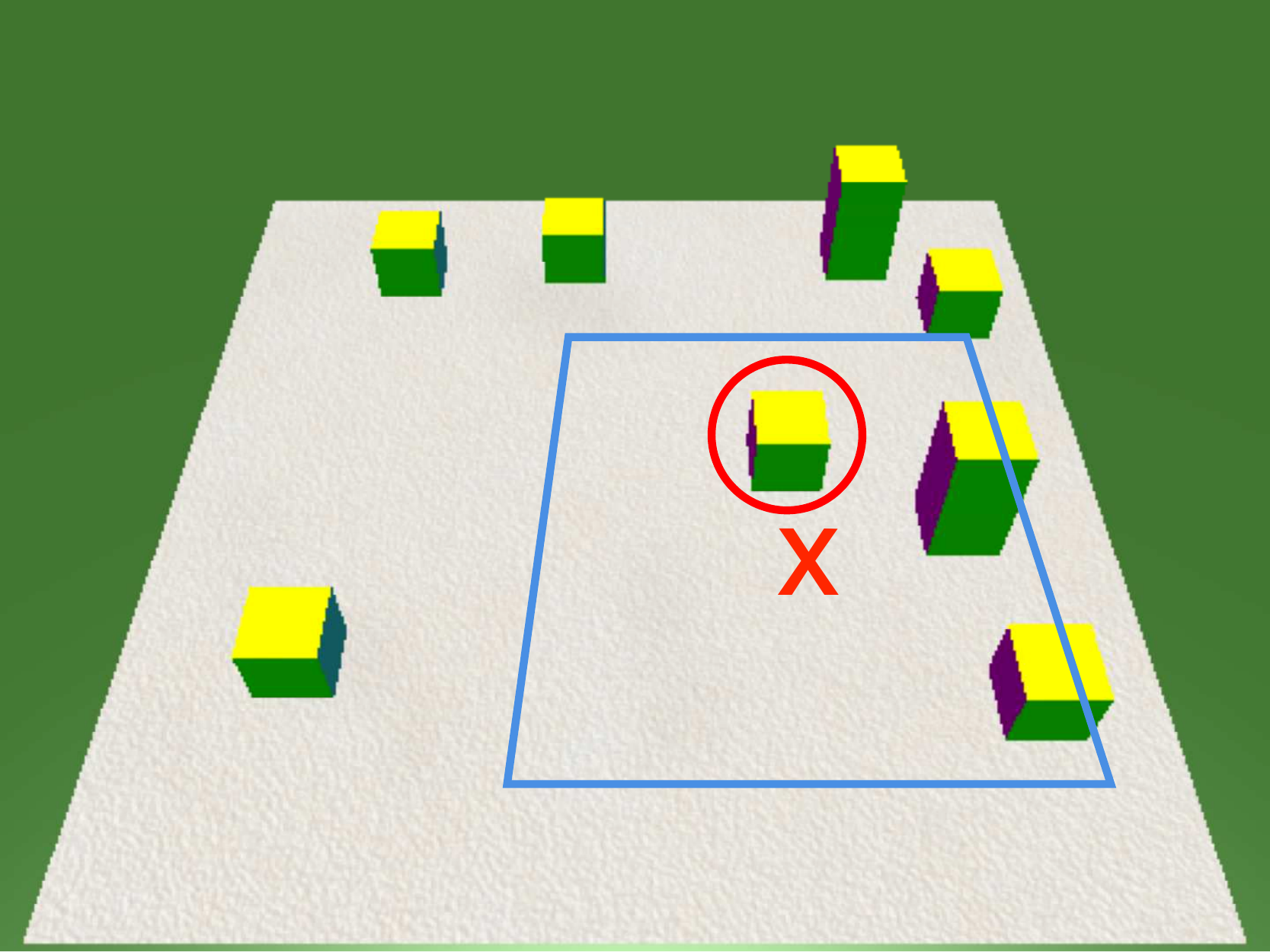}}
\caption{Benefit of Input-Specific Model Self-Generated Advice \newline (a) The \citeauthor{bisk2016natural} model would have made a prediction (`x') close to the true block (square). However, the advice region (blue) was incorrect (due to the true block being close to the edge of it) and this led to a significantly worse prediction (circle). \newline (b) In the input-specific self-generated advice model, the advice region (blue) is centered at the incorrect coordinate prediction (`x'), leading to the true source block being included and a correct prediction (circle).}
\label{fig:advanced_advice_gen}
\end{figure}

Our second approach for self-generated advice aims to improve on some of the shortcomings of the first approach. Previously, when generating the advice, we had decided on four coarse-grained regions, and trained a model to classify each input example into one of these regions. In many cases, the true coordinate lay close to the boundary of one of these regions, often resulting in the model predicting the wrong region when self-generating the advice. This incorrect prediction would lead to significantly worse performance (when compared to the model without advice) when running the end-to-end model from Section~\ref{sec:restrictive-advice}, as the advice was incorrect (remember that the model always follows our advice, and the true coordinate is not in the advice region due to the mistake). However, if we had instead chosen our regions to be centered at the true coordinate of each input example, it would be less likely that the model would make an incorrect region prediction (since a small error would still lead to the region containing the correct coordinate). Figure~\ref{fig:advanced_advice_gen} provides a visual explanation of this.

For this reason, we now introduce input-specific model self-generated advice. In this case, we run the \citeauthor{bisk2016natural} coordinate prediction model in two iterations. In the first iteration, we use the prediction to generate advice for a region (of the same size as in the case of 4 quadrants) centered at the predicted coordinate (see Figure~\ref{fig:advanced_advice_gen}b).\footnote{We make sure the advice region doesn't exceed the board boundaries.} In the second iteration, we feed in this generated advice just like Section~\ref{sec:restrictive-advice}. This model ($\mathtt{M9}$) achieves performance slightly worse than retry advice, and significantly better than \citeauthor{bisk2016natural}, all with no human effort.\footnote{Note that we must re-train the model from Section~\ref{sec:grounding} as there are now significantly more regions. The accuracy of that model is still 99.99\%, and the training procedure does not change.} Table~\ref{table:accuracyGenAdvice} shows the accuracy increase in predicting the advice now ($\mathtt{A3}$ vs $\mathtt{A1}$). It is unsurprising that this approach to self-generating advice performs better, as now the regions are more specific to each coordinate (so there is a higher probability that the true coordinate is actually in the predicted region - see Figure~\ref{fig:advanced_advice_gen}). 

We hypothesize that the performance improvements in self-generated advice happen since it is easier to predict the general region used to generate the advice rather than the specific coordinates. Previously, we have also shown the benefit of restrictive advice in improving overall coordinate prediction, so it is unsurprising that a high accuracy of advice generation leads to better overall performance. Due to this, we propose that future robot communication works take advantage of predicting and then using model self-generated advice in their end-to-end training procedure.

%% file: summary.tex
This paper takes a first step towards a stronger interaction between automated agents and their human operators, for physically grounded language understanding tasks. We focus on the popular blocks task and introduce the notion of advice, Natural Language hints provided by the human operator, correcting the model's predictions. We show that using four versions of this interactive advice driven protocol on an existing robot communication architecture, we can obtain significant performance improvements. The last method, model self-generated advice, shows the benefit of considering advice even when not designing an interactive protocol. Our future work focuses on further increasing the accuracy of the self-generated advice model, so we can achieve better performance with no human effort.

%% file: acknowledgements.tex
We thank the anonymous reviewers of this paper for all of their vital feedback. 

This work was partially supported by the Defense Advanced Research Projects Agency (DARPA) under the ASED program. Any opinions, findings and conclusions or recommendations expressed in this material are those of the author(s) and do not necessarily reflect the views of the DARPA.